\documentclass[a4paper,twocolumn,11pt,unpublished]{quantumarticle}
\pdfoutput=1

\usepackage[numbers,sort&compress]{natbib}
\usepackage[utf8]{inputenc}
\usepackage[english]{babel}
\usepackage[T1]{fontenc}
\usepackage{amsmath}
\usepackage{booktabs}
\usepackage{multirow}
\usepackage{graphicx}

\usepackage{bm}

\usepackage{caption}
\usepackage{subcaption}

\usepackage{hyperref}
\hypersetup{
    colorlinks=true,
    linkcolor=blue,      
    citecolor=blue,     
    filecolor=magenta,   
    urlcolor=blue        
}

\usepackage{tikz}
\usepackage{lipsum}

\newcommand{\upplus}{\raisebox{0.4ex}{\tiny$\bm{+}$}}
\newcommand{\semibold}[1]{{\textbf{\scalebox{0.9}[1.0]{#1}}}}
\newcommand{\algorithm}{\textsc{sprout}}

\newcommand{\mpl}{Max Planck Institute for the Science of Light, Erlangen, Germany}
\newcommand{\ligo}{LIGO Laboratory, California Institute of Technology, Pasadena, USA}
\newcommand{\jku}{ELLIS Unit Linz, Institute for Machine Learning, JKU Linz, Austria}
\newcommand{\tuebingen}{Machine Learning in Science Cluster, Department of Computer Science, Faculty of Science, University of Tuebingen, Germany}
\newcommand{\ta}{School of Physics and Astronomy, Tel Aviv University, Israel}
\newcommand{\emmi}{Emmi AI, Linz, Austria}

\begin{document}

\title{Neural surrogates for designing gravitational wave detectors}

\author{Carlos Ruiz-Gonzalez}%
\email{cruizgo@proton.me}
\affiliation{\mpl}
\affiliation{\jku}

\author{Sören Arlt}
\affiliation{\mpl}
\affiliation{\tuebingen}

\author{Sebastian Lehner}
\affiliation{\jku}

\author{Arturs Berzins}
\affiliation{\jku}

\author{Yehonathan Drori}%
\affiliation{\ligo}
\affiliation{\ta}

\author{Rana X Adhikari}
\affiliation{\ligo}

\author{Johannes Brandstetter}
\affiliation{\jku}
\affiliation{\emmi}

\author{Mario Krenn}
\email{mario.krenn@uni-tuebingen.de}
\affiliation{\mpl}
\affiliation{\tuebingen}

\maketitle

\begin{abstract}
Physics simulators are essential in science and engineering, enabling the analysis, control, and design of complex systems. In experimental sciences, they are increasingly used to automate experimental design, often via combinatorial search and optimization. However, as the setups grow more complex, the computational cost of traditional, CPU-based simulators becomes a major limitation. Here, we show how neural surrogate models can significantly reduce reliance on such slow simulators while preserving accuracy. Taking the design of interferometric gravitational wave detectors as a representative example, we train a neural network to surrogate the gravitational wave physics simulator Finesse, which was developed by the LIGO community. Despite that small changes in physical parameters can change the output by orders of magnitudes, the model rapidly predicts the quality and feasibility of candidate designs, allowing an efficient exploration of large design spaces. Our algorithm loops between training the surrogate, inverse designing new experiments, and verifying their properties with the slow simulator for further training. Assisted by auto-differentiation and GPU parallelism, our method proposes high-quality experiments much faster than direct optimization. Solutions that our algorithm finds within hours outperform designs that take five days for the optimizer to reach. Though shown in the context of gravitational wave detectors, our framework is broadly applicable to other domains where simulator bottlenecks hinder optimization and discovery.
\end{abstract}

\section{Introduction}

Computer simulations are ubiquitous in all branches of science and engineering, providing predictions and a deeper understanding of complex interacting systems \cite{metropolis1980history,
battimelli2020computer, dongarra2024co}. In experimental sciences, they are also used to optimize, control, and even design experimental setups. This goes beyond tuning a few parameters; novel algorithms have (partially) automated the conception of experiments, leading to new and often counterintuitive designs \cite{krenn2020computer}. In physics alone, there are successful examples in quantum physics \cite{krenn2016automated,knott2016search,o2019hybrid, melnikov2020setting,ruiz2023digital,maclellan2024end,lanore2024automated}, particle physics \cite{dorigo2023toward,rolla2023using,strong2024tomopt}, nano-optics \cite{molesky2018inverse,sapra2020chip,yang2023inverse}, microscopy \cite{nehme2020deepstorm3d,rodriguez2024automated} and gravitational wave physics \cite{krenn2023uifo}, to name a few.

The inverse design of experiments is extremely computationally intensive \cite{platt2018computational}, especially when combinatorial search is required. Therefore, the simulator speed is paramount to navigate the vast space of experimental designs. Fortunately, these high computational demands are mitigated by hardware improvements, such as GPUs \cite{dally2021evolution, silvano2025survey}, and high-performance tools, like auto-differentiation or just-in-time compilation. These advances can also benefit slower CPU-based software, which can be surrogated with neural networks \cite{asano2018optimization, ma2021deep, jiang2021deep, kudela2023deep, grady2023model, yin2023solving, peano2021rapid, schmidt2025endtoend}.

In this work, we leverage fast neural surrogate models to design interferometric gravitational wave detectors (GWDs). These sensitive experiments, able to detect distant cosmological events, can be simulated with the CPU-based software Finesse \cite{finesse}. This open-source simulator was developed by the LIGO collaboration, the first to detect gravitational waves in 2015 \cite{abbott2016observation}, and is the de facto standard to GWD design. In recent work  \cite{krenn2023uifo}, Finesse was used to automate the search of new interferometric detectors by optimizing a highly parametrized experimental ansatz \cite{krenn2023uifo}. However, as the Finesse simulator is neither differentiable nor GPU accelerated, gradient-based optimization becomes increasingly inefficient when the system grows.

To accelerate the design process, we present \algorithm, short of \textit{Surrogate Predictions for Reiterative Optimization with Update Training}. Our multi-step approach trains a surrogate model to predict the quality and feasibility of different GWDs, encoded as the parameters of a UIFO. First, we train the model with randomly parametrized designs, to then generate new data that, once verified, is used for further training. Iterating the design and training (see Figure \ref{fig:firstiterations}), we produce successive generations of models that efficiently explore the design space and improve the solutions. This iterative optimization process, also known as online optimization or active learning \cite{active2009, li2024survey}, is a successful strategy when simulations (or experiments) are expensive \cite{kandasamy2016gaussian, tran2018active, lookman2019active, van2024traversing}. 

Our approach does not replace the original simulator, as we still use it to verify the designs generated by the faster GPU-based differentiable surrogates. However, by rapidly exploring the space of possible designs, even if it is an approximation, we reduce the overall usage of the CPU software and still obtain high-quality designs. Further, since the verifications of the GPU-based optimization are independent of each other, we can parallelize the simulator calls arbitrarily, beyond what is possible in a sequential gradient-based optimization. Finally, when the size of the GWD increases, the growing number of tunable parameters and the longer simulation times make the numerical gradient approximations prohibitively slow. Neural surrogates excel in such a scenario, as backpropagation and the vast parallelization capabilities of GPUs allow us to optimize thousands of experiments at breakneck speed. We obtain high-quality designs in a fraction of the time, and, remarkably, even when we train the models with randomly parametrized designs, they find solutions far beyond anything we trained them with (see Figure \ref{fig:inverse_benefit}).

\begin{figure*}[t!]
  \centering
  \includegraphics[width=\textwidth]{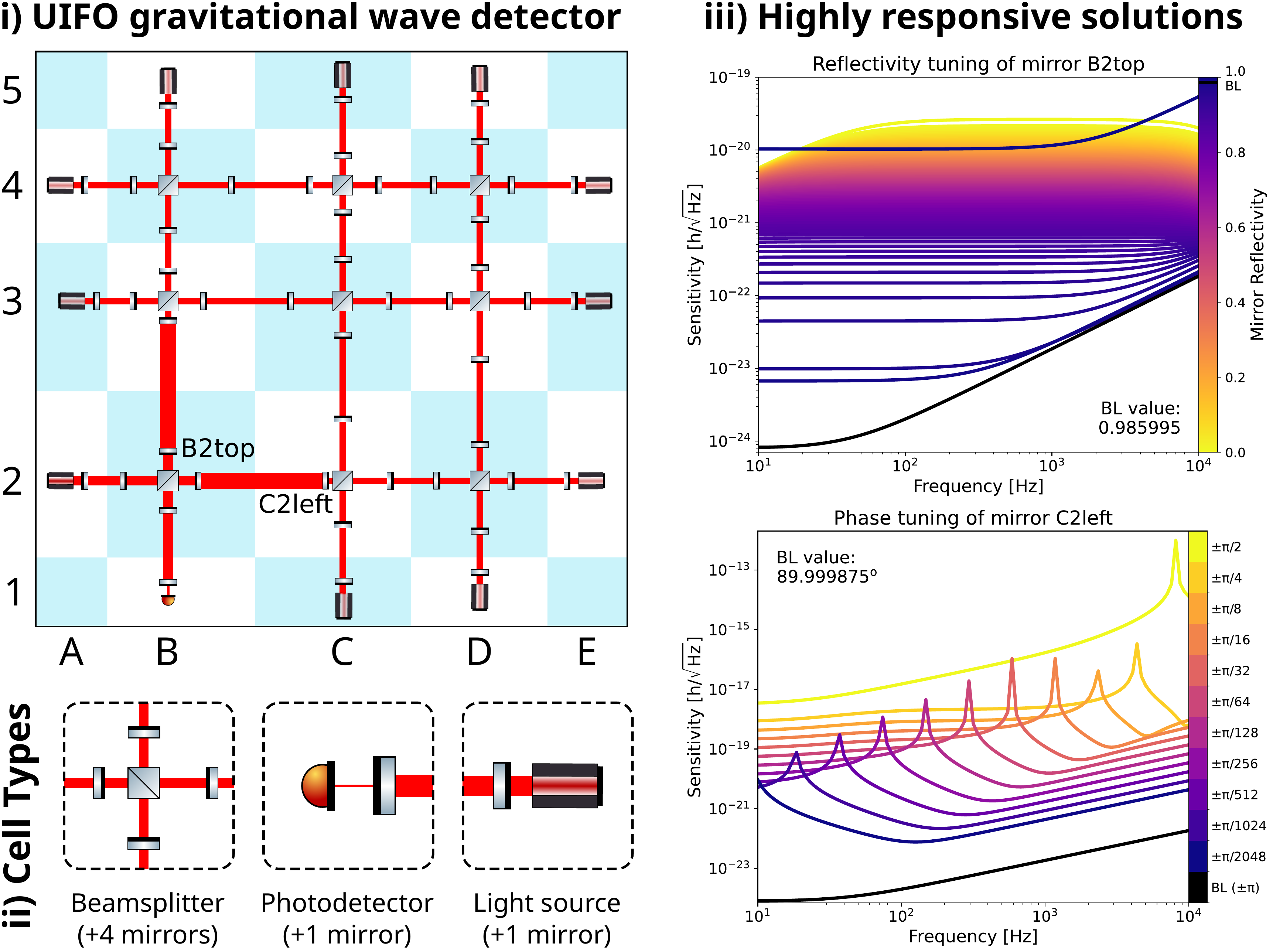}
  \caption{\textbf{A highly expressive ansatz: quasi-Universal InterFerOmeter (UIFO).} A large variety of gravitational wave detectors can be built by aligning optical elements in an irregular rectangular grid. This expressive parametrized template is referred to as quasi-Universal InterFerOmeter, or UIFO. \textbf{i)} The 3x3 UIFO, with 169 parameters, has 9 beamsplitter cells surrounded by light source cells and a single photodetector. By tuning the parameters of the grid's components, using only a small subset of components, we can recreate a simplified version of the GWD used by the LIGO collaboration. \textbf{ii)} In the depicted design and throughout this paper, we use beamsplitters, mirrors, photodetectors, and lasers. \textbf{iii)} The quality of a GWD is given by the smallest spatial deformation that it can detect for each gravitational wave frequency, represented by a sensitivity curve. As shown in the top plot, these curves can fluctuate in several orders of magnitude in response to a single parameter, like the reflectivity of the top mirror in cell B2. The change is even more drastic when tinkering with the left mirror in cell C2.}
  \label{fig:UIFO}
\end{figure*}

\subsection{Related works}
\hspace{\parindent}\textbf{Surrogates for inverse design.} When the computational cost of traditional simulators becomes unbearable, (neural) surrogate models are efficient, often differentiable, alternatives \cite{forrester2008engineering}. They are widely used for inverse design tasks in multiple fields, including engineering \cite{testolina2019antenna, chang2020learning, yang2023turbine}, chemistry \cite{ma2018deep, wang2021inverse, xue2016accelerated, shen2021deep, Jaouni2024}, or physics \cite{edelen2020accelerator, shirobokov2020black, peano2021rapid, schmidt2025endtoend, wozniak2025endtoend}, being particularly popular in photonics \cite{liu2018nanostructures,asano2018optimization,ma2021deep,jiang2021deep,kudela2023deep}. Most of these surrogates predict one or a few values that summarize the designs' quality. In contrast, we predict an array of properties that indicate the quality as well as the hardware requirements to build the design. These predicted properties are incredibly responsive to the input parameters, and can fluctuate up to 10 orders of magnitude when tuning a single experimental parameter (see Figure \ref{fig:UIFO}). We did not find such sensitive systems in the surrogate literature and took inspiration from other domains \cite{fourierfeatures} to face the challenge (details on Section \ref{sec:methods}). 

\textbf{Neural networks in physics.} Recent advances in machine learning are especially aimed at the physical sciences. Physics-informed neural networks (PINN) \cite{raissi2019physics, karniadakis2021piml}, or neural operators \cite{lu2021deeponet,azizzadenesheli2024neural}, are notable examples, widely used as PDE surrogates in fluid dynamics  \cite{mahmoudabadbozchelou2022nn,alkin2024universal}, among other domains \cite{oommen2024materials, cranganore2025einsteinfields}. However, these novel approaches are not suited to our task: predicting the experiment's sensitivity and the light powers applied on the elements, based on the design parameters. For the PINNs, we lack simple physical laws to constraint the outputs. On the other hand, neural operators surrogate functions which can be evaluated at any discretization of a continuous input space. Here we have a finite set of input properties, from a finite number of optical parameters, and a fixed number of output properties. Thus, these tools offer us no advantage.

\textbf{Pool-based active learning.} Supervised learning requires costly labeled data. Thus, active learning strategies must carefully select which data to label, from a pool of unlabeled data points \cite{active2009}. There exist theoretical results about the strengths and limitations of several approaches \cite{dasgupta2004greedy, hanneke2008bound, ganti2012upal, gonen2013efficient}, but their assumptions rarely hold when training deep neural networks. Therefore, the field relies mostly on heuristics and empirical evidence for particular applications \cite{ross2011reduction, lakshminarayanan2017simple, wang2017cost, sener2018active}. Bayesian methods are also popular for small systems \cite{roy2001toward, kapoor2007active, gal2016dropout,gal2017deepbayes}, but they do not scale well. In this work, we verify the designs obtained after a noisy optimization process of the surrogate simulator, pushing the model to generalize beyond its training dataset. We repeat this procedure for 5 rounds (see Figure \ref{fig:activelearning}). Our strategy resembles previous work on active learning \cite{pestourie2020active, singh2025active} and on adversarial training \cite{goodfellow2014explaining}, where gradient-based optimization generates adversarial examples that are mislabeled by neural  \cite{kurakin2017adversarial,madry2018towards}. Retraining the networks with such examples increases their robustness and improves their predictions in high-quality design spaces. 

\section{Methods}\label{sec:methods}
This section describes how to design GWD using the \textit{quasi-Universal InterFerOmeter} (UIFO), a highly overparametrized experimental setup introduced in previous work \cite{krenn2023uifo}, which is used as a highly expressive ansatz. We also outline the surrogate models' architecture and training strategy, the building blocks of the \algorithm\: algorithm.

\subsection{Gravitational wave detectors}\label{parametrizedGWD}

Gravitational waves are perturbations of the spacetime caused by the motion of masses. They were predicted by Einstein's general relativity in 1916 \cite{einstein1916naherungsweise} and directly observed for the first time in 2015 by the LIGO collaboration \cite{abbott2016observation}. Feeble as they are, we can only detect the waves originating from massively energetic astrophysical phenomena, such as supernovae or the merging of black holes. To do so, experimentalists use some of the most sensitive devices ever created by humans -- gravitational wave detectors -- massive facilities whose components require extremely precise tuning. Yet, as shown in previous work \cite{krenn2023uifo}, the space of GWD experiments is mostly unexplored. Many possible designs, including the LIGO detector, can be described within the UIFO framework, a rectangular lattice of parametrized optical elements. Using an expressive ansatz is a powerful tool to design new experiments, also in other fields \cite{sim2019expressibility, ruiz2023digital, rodriguez2024automated}, as it relaxes the optimization problem from (partially) discrete to purely continuous. The price we pay is to optimize larger systems of tunable elements, likely redundant, which contain previously known baselines and new designs able to surpass them.

As shown in Figure \ref{fig:UIFO}, for a given choice of UIFO parameters, the detector will have a certain strain sensitivity for each possible frequency of a gravitational wave, which we refer to as \textit{sensitivity curve}. The lower the sensitivity, the smaller space fluctuations it can detect. Additionally, it is necessary to limit the optical power in each component, otherwise their performance degrades. The same applies to the photodetector, which is orders of magnitude more sensitive. These and other constraints are summarized in Table \ref{tab:bounds}.

The sensitivity and power values define the quality and viability of the designs and can be computed with Finesse, the CPU-based simulator we want to surrogate. As shown in Figure \ref{fig:UIFO}iii), the quality of the interferometer can fluctuate by multiple orders of magnitude when tuning even a single parameter. Furthermore, some of the plotted sensitivity curves could not be implemented with current hardware, as the light power at several components surpass the forementioned limitations.

In this project, we focus on the continuous optimization of the parametrized ansatz UIFO and how the benefit of the surrogates increases with the size of the GWD. Therefore, to facilitate the comparison between designs of different sizes, we fix all the discrete design variables: the grid size, the beam splitters' orientations, and the location of the photodetector. The reduced size of our setups also facilitates the experiments. For more practical applications, we would emulate larger designs with different configurations that include even more sensitive designs than the LIGO detector. For further details on the UIFO parametrization, see Appendix \ref{app:characterization}.

\begin{table}[h!]
\centering
\begin{tabular}{@{}lll@{}}
\toprule
\textbf{Parameter}        & \textbf{Min} & \textbf{Max} \\ \midrule
Optical path length             & 1m           & 4km          \\
Optical loss per element& 5ppm         &              \\
Optical transmission     & 15ppm        &              \\
Reflected optical power   &              & 3.5MW        \\
Transmitted optical power &              & 2kW          \\
Power in photodetector    &              & 10mW         \\ \bottomrule
\end{tabular}
\caption{\textbf{Ranges of the physical parameters of the UIFO designs.} The loss/transmission of the optical elements can be bounded before running the simulations. Same for the optical path length, which ensures that the design remains within the 4km that LIGO design measures. The powers at the optical elements, computed by the simulator, are enforced via penalties. The lower bound for the optical loss, 5ppm, is a rough estimate of what is possible with current technology. We assume the lowest possible loss for all components, as larger values would only worsen the performance. Further details in Appendix \ref{app:characterization}.}
\label{tab:bounds}
\end{table}

\begin{figure*}[t]
    \centering
    
\begin{subfigure}[]{\textwidth}
    \centering
    \includegraphics[height=47mm]{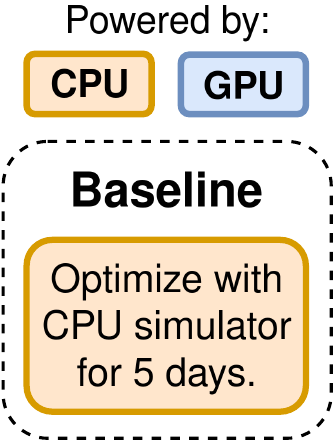}
    \hspace{5mm}
    \includegraphics[height=47mm]{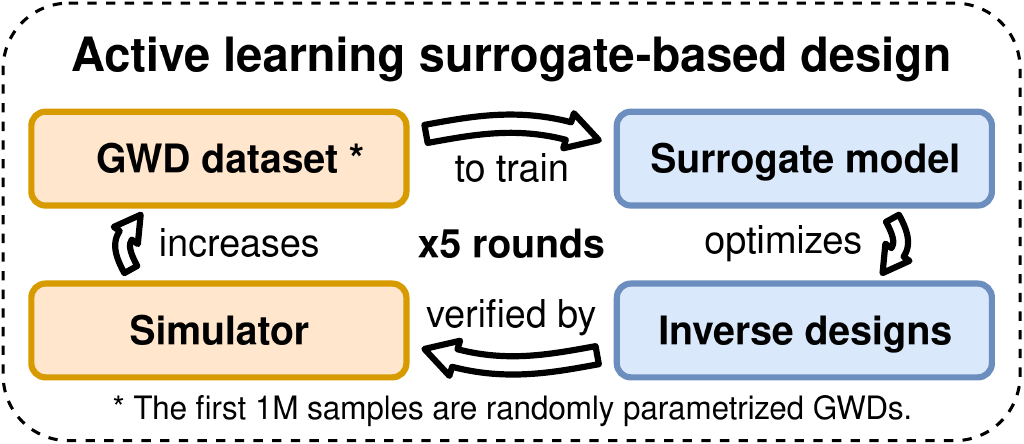}
    
\caption{Our active learning approach uses GPU units to reduce the usage of Finesse, a computationally expensive simulator.}
\vspace{3mm}
\label{fig:activelearning}
\end{subfigure}

\begin{subfigure}[]{\textwidth}
    \includegraphics[width=1\textwidth]{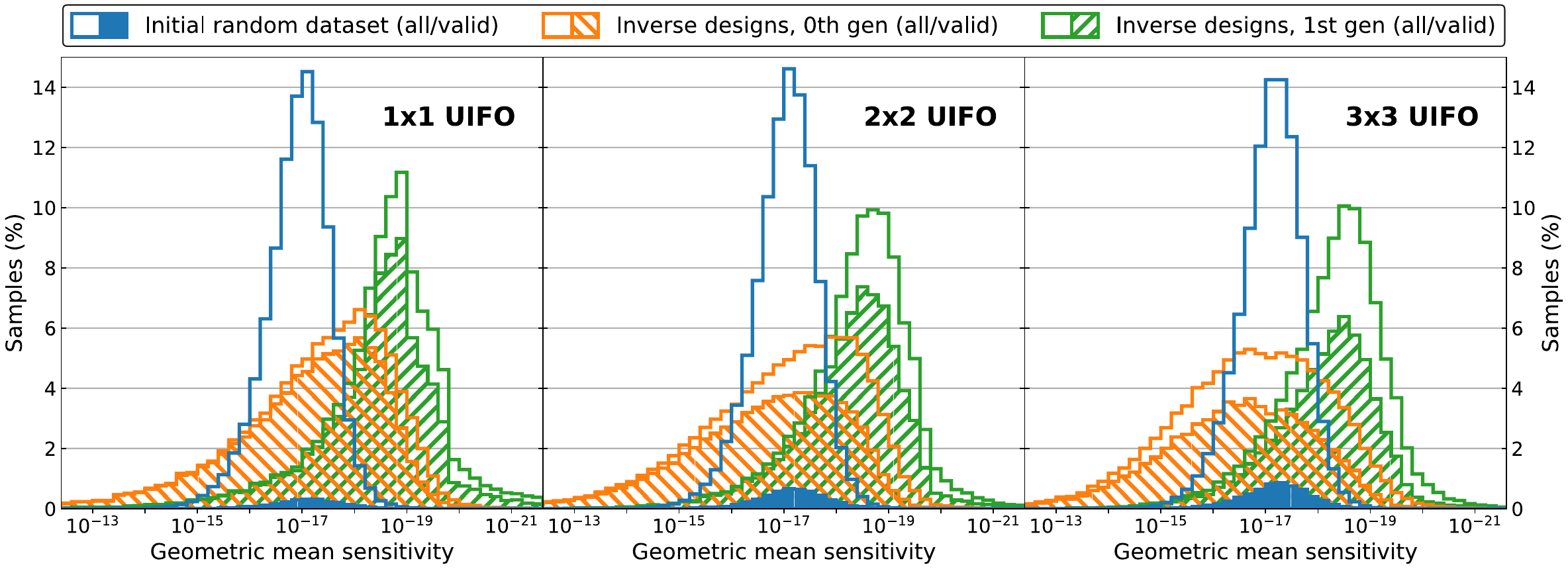}
\caption{The designs' quality (i.e. low sensitivity) surpasses the model's training data. Only valid samples meet hardware constrains.}
\vspace{3mm}
\label{fig:inverse_benefit}
\end{subfigure}

\begin{subfigure}[]{\textwidth}
    \centering
    \resizebox{\linewidth}{!}{\begin{tabular}{|c|c|c|c|c|c|c|}
\hline
\textbf{Experiment size}  & \multicolumn{2}{|c|}{\textbf{1x1 (29 parameters)}}& \multicolumn{2}{|c|}{\textbf{2x2 (85 parameters)}}& \multicolumn{2}{|c|}{\textbf{3x3 (169 parameters)}}\\ \hline
\textbf{Design strategy}  & \textbf{Simulator}           & \textbf{Surrogate} & \textbf{Simulator}           & \textbf{Surrogate} & \textbf{Simulator}            & \textbf{Surrogate}  \\ \hline
\textbf{Processor unit}   & CPU only            & CPU+GPU   & CPU only            & CPU+GPU   & CPU only             & CPU+GPU    \\ \hline
\textbf{Time until 1e-20} & 1h10'               & 14'+40'   & 6h05'               & 41'+1h09'  & 32h15'               & 1h23'+2h01' \\ \hline
\begin{tabular}[c]{@{}c@{}} \textbf{Simulation calls} \vspace{-1mm}\\ \textbf{to reach 1e-20}\end{tabular} & 5.5M & 2.2M & 9.6M & 2.2M & 25.6M & 2.2M \\ \hline
\end{tabular}}
\vspace{1mm}
\caption{For the same quality, the surrogate-based design (+ random sampling) takes much less time than only using the simulator.}
\end{subfigure}
\caption{\textbf{Iterative inverse design with surrogate models.}}
\label{fig:firstiterations}
\end{figure*}

\subsection{Design of experiments} \label{sec:design}
Whether we simulate the UIFO with Finesse or rely on a neural surrogate, the strategy to design novel GWD is the same. Having fixed the discrete design choices, we optimize the continuous parameters of the UIFO, our template for GWD, to minimize the geometric mean of the strain sensitivity.

Furthermore, to guarantee the feasibility of the designs, they must meet the hardware constraints described in Table \ref{tab:bounds}, which can be divided into two types. First, we have the maximum powers that the detectors, mirrors, and beam splitters can withstand. These are output constraints that are calculated by the simulator. We enforce them by adding penalties to the loss, using the differentiable function
\begin{equation}
\label{eq:penalties}
    \text{Penalty} \equiv (\alpha +\beta(x-x_\text{th})) \sigma(\gamma (x-x_\text{th})),
\end{equation}
which approximates a step function with a step size $\alpha$ and a final slope of $\beta$, for $x>x_\text{th}$. The stiffness of the step is indicated with $\gamma$, $\sigma$ is a sigmoid function, $x$ is the (log) optical power applied to a given optical element, and $x_\text{th}$ is the maximum power that the element can tolerate. The values of $x_\text{th}$ depend on the optical element and whether the power is reflected or transmitted (see Table \ref{tab:bounds}). We set the values of $\{\alpha, \beta, \gamma\}$ to $\{10, 1, 1000\}$.

In contrast, the constraints on element properties are enforced by restricting the input parameter space, for example, by limiting the reflectivity and losses of the mirrors and beam splitters. To guarantee spatial consistency, the distances between rows and columns of beamsplitters are described as fractions of a total larger distance, which must remain below 4 km. Finally, the positions of the different mirrors are also described by fractions (see Appendix \ref{app:characterization}).

The final loss is the mean of the log sensitivity for $N$ frequency values $f_i$ (i.e.~the log of the geometric mean) plus the described penalties,
\begin{equation}\label{eq:loss}
    \mathcal{L}=\frac{1}{N}\sum_{i=1}^N\log(\text{S}(f_i)) + \text{Penalties}.
\end{equation}
The set of frequencies are 101 logarithmically spaced values between 1Hz and 10kHz. 

To minimize the loss, we use the Adam optimizer \cite{adam} with learning rate of $\lambda=0.1$. Moreover, we add annealed Gaussian noise, $\mathcal{N}(0,2\epsilon^{t}\lambda)$, to the input update, a well-known strategy to optimize highly non-convex functions \cite{hinton2002stochastic, welling2011bayesian, gulcehre2016noisy}. With $\epsilon<1$, the noise decreases exponentially at every step $t$.

\begin{figure*}[t!]
  \centering
  \includegraphics[width=0.95\textwidth]{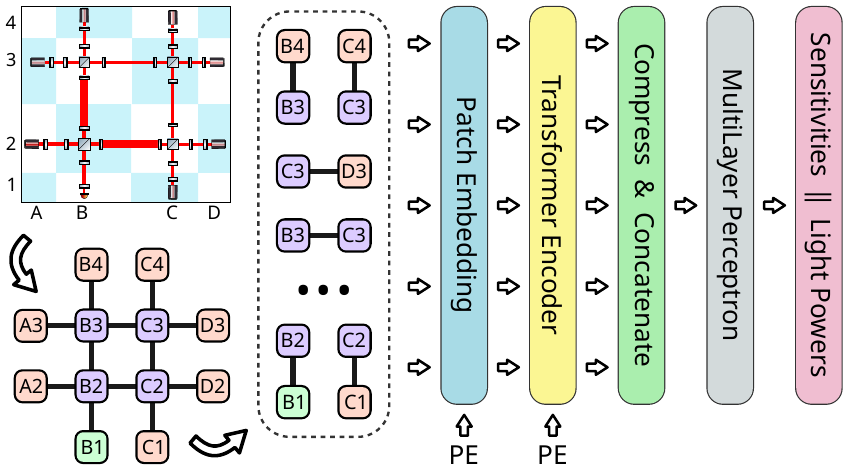}
  \caption{\textbf{Overview of the surrogate model's architecture.} After extracting the patches from the grid-based GWD, we transform them into (larger) token representations, which are then processed by a standard transformer encoder. To turn the tokens into the final output, we compress them with a learned projection and concatenate them. The resulting array, larger than the output by design, goes through a final MLP and, after a sigmoid, is turned into the output: the values of the sensitivity curve concatenated with the light powers at each optical element.}
  \label{fig:architecture}
\end{figure*}

\subsection{Surrogate models}
In order to find new GWD designs, our neural networks must be able to simulate them, at least approximately. We want our surrogates to predict, based on the input parameters of the UIFO, the final sensitivity curve of the designs (their quality) and the light powers that are applied to each optical component (their feasibility). To process the grid-like structure of the UIFO template, we divide the interferometer into overlapping subgraphs, each with a different set of parametrized elements, and turn them into patches, which are then processed by a transformer encoder and mapped into the output space. Our architecture is heavily inspired by models such as Vision Transformers \cite{visiontransformers}, MLP-Mixers \cite{mixers4images}, and other works that structure the input data into patches/tokens, \cite{mixers4graphs}. It has three main parts:\begin{enumerate}
    \item the patch extraction from the GWD and their embedding into tokens,
    \item the processing of the tokens with a standard transformer encoder of 8 layers, and 
    \item the decoding of the processed tokens into the final output of sensitivities and powers.
\end{enumerate} 

\textbf{Turning GWD into patches.} To transform a GWD into a set of patches, we treat the UIFO grid as an attributed graph, i.e. a graph whose nodes and edges have several weights. The nodes are the beam splitters, the light sources, and the detectors, each with a different set of associated parameters. Between the nodes, we always find two mirrors, whose properties and relative position in the optical paths are the attributes of the edges that connect the nodes. Grouping the parameters of an edge and the two adjacent nodes, we define a patch. Since the nodes have different numbers of attributes, we pad the patches with zeroes up to length 12. As shown in Figure \ref{fig:architecture}, the patches overlap in the beam splitters, as their respective nodes have degree 4.

To the padded patches, we add a positional embedding (PE) and apply the trainable Fourier feature map, that is, a linear layer followed by a sine activation \cite{fourierfeatures}. This mapping helps to overcome the spectral bias of neural networks towards low-frequency functions \cite{spectralbias,frequencybias}, which can limit the capacity of our models to predict the properties of high-quality outliers. The projection of the Fourier map increases the embedding dimension of the patches to 192. It is followed by a linear transformation.

\textbf{Transformer architecture.} After adding a positional embedding to the tokens, we pass them to the transformer encoder \cite{transformers}. We use eight standard transformer layers with the pre-layer normalization convention \cite{layernorm}, and a multilayer perceptron (MLP) with hidden dimension 384, twice the size of the tokens. We chose the transformer architecture, as it has shown great success processing multiple types of data for a wide variety of tasks \cite{lee2019set, devlin2019bert, schwaller2019molecular, alphafold, zhou2021informer, chen2021decision, janner2021offline, liu2021Swin, han2022survey}. 

\textbf{Towards the final output.} After the transformer encoder, we apply a layer norm and reduce the tokens' size with a linear projection, followed by a GELU. Concatenating the tokens, we obtain a final vector that, after a norm layer and an MLP, takes the shape of the output. We enforce normalization by applying a sigmoid to the output: the array of sensitivities and the light powers.

A summary of the architecture is shown in Figure \ref{fig:architecture}. The implementation details are described in Appendix \ref{app:architecture}, and all the code can be found in the repository\footnote{\href{https://github.com/artificial-scientist-lab/SPROUT}{https://github.com/artificial-scientist-lab/SPROUT}}.

The training datasets consist of pairs of input parameters and output sensitivities and powers. The input parameters are linearly normalized between 0 and 1, adjusting the rescaling to the different types of physical parameters. On the other hand, since the values of sensitivities and optical powers range across several orders of magnitude, we take their logarithm value for training. For further details on data preprocessing, see Appendix \ref{app:datasets}.

\begin{figure*}[t]
  \centering
  \includegraphics[width=\textwidth]{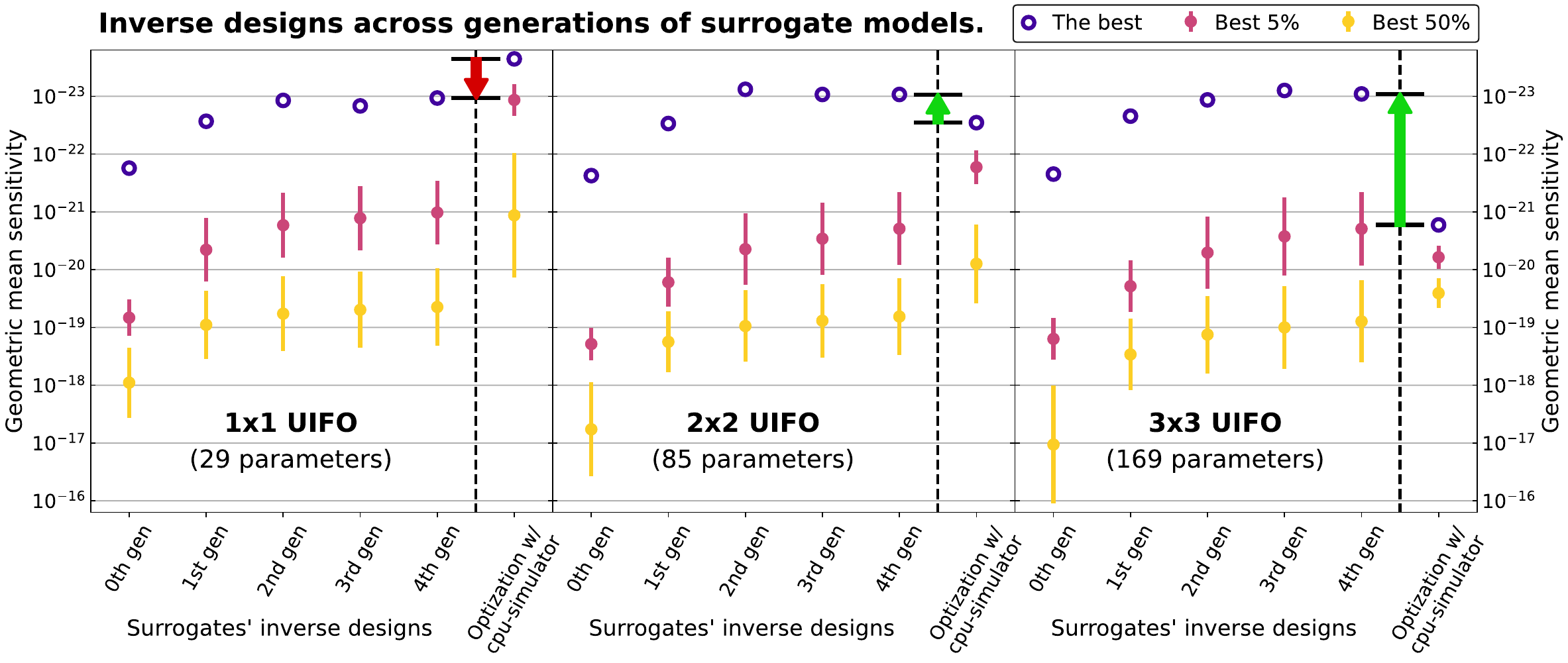}
  \caption{\textbf{Evolution of the inverse designs' quality after (re)training the surrogate model and comparison with direct optimization.} After training the surrogate model with random data we use it to produce hundreds of thousands of novel solutions via gradient-based optimization. Once the designs are verified, they are added to the previous training dataset, to finetune the model. At every generation of finetuned models the training data adds 1 million of new samples. The plotted sensitivities, which improve after every generation, come from the last step of the optimization process of the 200K inverse-designed samples. The optimizations with Finesse, detailed in Figure \ref{fig:direct_optimization}, were limited to 256 samples per grid size and ran for 5 days. The benefits of the surrogate models are specially prominent in the larger systems.}
  \label{fig:surrogate_designs}
\end{figure*}

\subsection{Active learning strategy}\label{sec:activelearning}

To train the models, we use the Adam optimizer with weight decay and a plateau learning rate schedule. In the first training phase, the model learns from 1 million randomly parametrized designs.\footnote{Additional 10\% are produced for testing.} With the trained models we run the Adam optimization process described in section \ref{sec:design}, adding annealed noise with $\epsilon=0.999$. Running 200,000 optimization trajectories for 16,384 steps on multiple GPU nodes, we store 5 designs for each trajectory\footnote{We store the designs at (approximately) logarithmically spaced steps: 6, 48, 337, 2352, and 16384.} producing one million new training samples each round (+10\% for testing). After verifying the designs  -- in parallel -- with the CPU-simulator Finesse, we use the old and the new samples, 2 million in total, to retrain the model. The total amount of training samples increases by 1 million at every training round, which are added to the training datasets in groups of 50K to stabilize training \cite{bengio2009curriculum}. The neural networks are initialized with the weights of the previous training phase.

It is noteworthy that the quality of the first training dataset, randomly produced, is always very low. Less than 5\% are valid, as most of them apply excessive light power in at least one optical component, usually the fragile detector. Yet, the first trained model produces valid designs much more sensitive than any sample it was trained with (see Figure \ref{fig:firstiterations}). That happens for all system sizes. After further training, the model produces increasingly better samples by learning from its previous designs, a virtuous cycle of self-improvement (see Figure \ref{fig:surrogate_designs}).

\section{Experiments}
To benchmark our surrogate-based algorithm, \algorithm, we compare it with the previous baseline: direct optimization with Finesse, the CPU-based simulator. Table \ref{tab:massivetable} summarizes the most relevant feats of both approaches, while Figure \ref{fig:surrogate_designs} shows the evolution of the surrogates' designs across several iterations of (re-) training and designing, comparing them with the CPU-powered optimization process. We also detail the resources employed for both approaches.

\subsection{CPU-based optimization baseline}\label{sec:baseline}
The original approach we want to surpass is the gradient-based optimization with the CPU-simulator Finesse. Therefore, we optimize 256 randomly parametrized UIFO samples with the Adam algorithm, adding annealed noise of $\epsilon=0.999$, just as with the surrogates. The main difference from surrogate optimization is the need to compute the gradients with finite differences, a much slower procedure than automatic differentiation. We ran the optimization for 5 days and only the smallest UIFO with 29 parameters was able to reach more than 10 thousand optimization steps.

The final designs' sensitivities optimized with Finesse are shown in Figure \ref{fig:surrogate_designs} for the 3 UIFO sizes, together with the designs of the surrogate models. The complete evolution of the CPU-based optimization over the 5 days is shown in Figure \ref{fig:direct_optimization}.

\begin{table*}[t!]
\centering
\begin{tabular}{|c|cc|cc|cc|}
\hline
\semibold{UIFO grid size} &
  \multicolumn{2}{c|}{\semibold{1x1 (29 parameters)}} &
  \multicolumn{2}{c|}{\semibold{2x2 (85 parameters)}} &
  \multicolumn{2}{c|}{\semibold{3x3 (169 parameters)}} \\ \hline
\semibold{Design tool} &
  \multicolumn{1}{c|}{\semibold{Simulator}} &
  \semibold{Surrogate} &
  \multicolumn{1}{c|}{\semibold{Simulator}} &
  \semibold{Surrogate} &
  \multicolumn{1}{c|}{\semibold{Simulator}} &
  \semibold{Surrogate} \\ \hline
\semibold{Simulator calls}  & \multicolumn{1}{c|}{560.6M}       & 6.6M     & \multicolumn{1}{c|}{189.3M}       & 6.6M     & \multicolumn{1}{c|}{95.2M}    & 6.6M         \\ \hline
\semibold{Final samples}    & \multicolumn{1}{c|}{256}    & 1.1M     & \multicolumn{1}{c|}{256}    & 1.1M     & \multicolumn{1}{c|}{256}      & 1.1M         \\ \hline
\semibold{Steps / sample}   & \multicolumn{1}{c|}{73001}       & 16384    & \multicolumn{1}{c|}{8600}       & 16384    & \multicolumn{1}{c|}{2189}     & 16384        \\ \hline
\semibold{Best sensitivity} & \multicolumn{1}{c|}{\underline{2.25e-24}}       &  1.07e-23& \multicolumn{1}{c|}{2.85e-21}       & \underline{7.56e-24}& \multicolumn{1}{c|}{1.68e-21} & \underline{7.92e-24}     \\ \hline
\semibold{Time spent}&
  \multicolumn{1}{c|}{\semibold{\small CPU}}&
  \semibold{\small CPU\upplus GPU}&
  \multicolumn{1}{c|}{\semibold{\small CPU}}&
  \semibold{\small CPU\upplus GPU}&
  \multicolumn{1}{c|}{\semibold{\small CPU}}&
  \semibold{\small CPU\upplus GPU} \\ \hline
\semibold{Until 1e-20}      & \multicolumn{1}{c|}{1h10'} &   14'\upplus40'    & \multicolumn{1}{c|}{6h05'}       &   41'\upplus1h09'      & \multicolumn{1}{c|}{32h15'}   & 1h23'\upplus2h01'    \\ \hline
\semibold{Until 1e-22}      & \multicolumn{1}{c|}{5h09'} &  28'\upplus1h44'   & \multicolumn{1}{c|}{90h13'}       &   1h23'\upplus2h56'   & \multicolumn{1}{c|}{Never}    & 2h46'\upplus4h37'  \\ \hline
\semibold{Total}& \multicolumn{1}{c|}{5 days} &     1h24'\upplus7h36'     & \multicolumn{1}{c|}{5 days} &   4h10'\upplus11h27'   & \multicolumn{1}{c|}{5 days}   & 8h18'\upplus18h6' \\ \hline
\end{tabular}
\caption{\textbf{Time, quality, and resource comparison of the 2 design strategies for 3 sizes of UIFO.} Our surrogate-based approach drastically reduced the simulator calls to achieve the same design quality. The CPU-powered optimization ran for 5 days (fixed) in 8 units of \textit{Xeon Gold 6130}, with 32 CPU cores each. With the same CPU resources, the verification of the 6.6M samples (5 rounds of inverse design + starting random dataset) takes only a few hours. The additional GPU time comes from the training time, which increased at every round with the size of the dataset (see Table \ref{tab:trainingtimes}), together with the inverse design, which took (approximately) 16, 40, and 85 minutes, for the 3 respective sizes. The inverse design was distributed across 11 GPU cores from multiple \textit{AMD Instinct MI300A APU} units, each optimizing 20K samples in parallel. Details on the computational resources in Appendix \ref{app:resources}.}
\label{tab:massivetable}
\end{table*}

\subsection{The SPROUT algorithm}
\algorithm\: overcomes the limitations of the CPU simulator by delegating part of the task to powerful and parallelizable GPU nodes. At each round, we train the model with a GPU node from \textit{AMD Instinct MI300A APU}, and for inverse design we use 11 of such nodes. We optimize the designs in batches of 20000, for all UIFO sizes, minimizing the loss from equation \ref{eq:loss}, as described in Section \ref{sec:activelearning}.

Storing five optimized samples for every optimization trajectory, we have 1 million of new input samples for training (+10\% for testing). To verify them, we run Finesse in parallel jobs across 256 CPU cores from several \textit{Xeon Gold 6130}. That is the same hardware used for the optimization with Finesse described in Section \ref{sec:baseline}.

To ensure the reliability of our approach, we applied \algorithm\: twice for each size of the UIFO. Therefore, the models are initially trained with the same random dataset but, due to the different random seed, they present minor differences in the training process that also affect the design (see Appendix \ref{app:resources}). Nonetheless, the overall quality is usually very similar. The inverse designs shown in Figure \ref{fig:surrogate_designs} are the aggregated data of both experiments, while the times in Table \ref{tab:massivetable} are averaged from the two processes.

\subsection{Results}\label{sec:results}
As shown in Table \ref{tab:massivetable}, the advantage of the surrogate models for inverse design increases with the size of the experiment. For the largest UIFO, the designs from the surrogate model improve the sensitivities by more than an order-of-magnitude. The increasingly better designs presented in Figure \ref{fig:surrogate_designs} show how the models are able to generalize beyond the data they are trained on. This is particularly noticeable at the first iterations (see Figure \ref{fig:inverse_benefit}). 

The key behind the surrogates' performance is to utilize the GPU and the differentiability of the neural nets, being able to simulate many more experiments in a fraction of the time than required by the CPU-based simulator Finesse. To give a loose comparison, for the larger 3x3 UIFO, an optimization step takes less than 0.3 seconds on a \textit{AMD Instinct MI300A APU} GPU core for 20,000 samples. A single evaluation with Finesse takes around 1.2 seconds on a \textit{Xeon Gold 6130} CPU node. Needing 169 additional simulator calls to compute the gradient, every optimization step can take more than 3 minutes. Therefore, as the size of the optimized GWD grows, so do the benefits of the surrogate models.

\section{Discussion}
We introduced \algorithm\:, a surrogate-based approach to accelerate the inverse design of gravitational wave detectors. Our algorithm navigates the space of experimental designs by iteratively training neural surrogates, optimizing designs, and selectively verifying candidates with Finesse, the non-differentiable, CPU-based simulator that we are surrogating. While tested with a limited set of experimental tools, the surrogates exhibited outstanding generalization capabilities, as they designed experiments of much higher quality than those used for training. Given the drastic fluctuations in the GWD's properties when modifying the input parameters, the feat is even more surprising.

By exploiting differentiability and GPU-based parallelization, the benefits of our neural network approach rise with the system size, as finite-difference methods become increasingly expensive. This approach transcends our field, as it can accelerate any design task that relies on non-differentiable, CPU-based simulators. Moreover, the patch-based encoding of experimental designs is a simple yet versatile approach, suited to all types of modular setups. Finally, our active learning strategy needs little domain expertise, only clear goals and constraints for the models to pursue.

With the positive results of our algorithm, there is still room for improvement. First, our inverse designs use a small set of optical elements and a few simple topologies. Constraining the task, we can compare the performance of different methods in a well-known arena, but we are limited to rediscovery and prone to stagnation. To reach SOTA detectors, we need other components and topologies. That leads us to the second point: our neural networks could be more flexible. We trained our neural networks on fixed discrete choices, but better models could pick different components, optimize the topology of experimental setups, or even predict the detector's sensitivity for arbitrary wave frequencies. To meet these ambitious goals, our architectures must map between different input and output spaces, and we need richer training data that encodes every possible design choice. These modifications might lead to general changes in the \algorithm\: algorithm, our third point. As the search space grows in dimensions and complexity, we might need more efficient training and exploration strategies. Some options include limiting the training time, intelligent sample weighting to favor specific designs' qualities, or implementing curriculum training with increasingly complex designs. Ensemble methods can foster exploration by aggregating predictions across diverse surrogate models, and diffusion-based design could be a powerful asset to explore different topologies.

To conclude, \algorithm\: is a promising tool for inverse design, whose principles and benefits apply beyond the search for GWD. Having showcased its potential in a constrained task, this work paves the way for further advances that lead to novel and more powerful experiments.

\section*{Acknowledgements}
The authors thank Daniel A. Brown and the members of the LIGO Scientific Collaboration Advanced Interferometer Configurations working group for many valuable conversations regarding Finesse and interferometers. M.K. acknowledges support by the European Research Council (ERC) under the European Union’s Horizon Europe research and innovation programme (ERC-2024-STG, 101165179, ArtDisQ), and by the German Research Foundation DFG (EXC 2064/1, Project 390727645).
C.R.G. acknowledges travel support from the European Union’s Horizon 2020 research and
innovation programme under ELISE Grant Agreement No 951847.
R.X.A acknowledges support by the Woodnext Foundation
and the John Templeton Foundation.
This work was made possible through the support of Grant 63405 from the John Templeton Foundation. 
The opinions expressed in this publication are those of the author(s) and do not necessarily reflect the views of the John Templeton Foundation.

\bibliographystyle{unsrtnat}
\bibliography{main.bib,intro.bib}

\onecolumn
\newpage
\appendix
\section{UIFO characterization}\label{app:characterization}
\subsection{Consistent distances}
\begin{figure}[h!]
    \centering
    \includegraphics[width=0.6\linewidth]{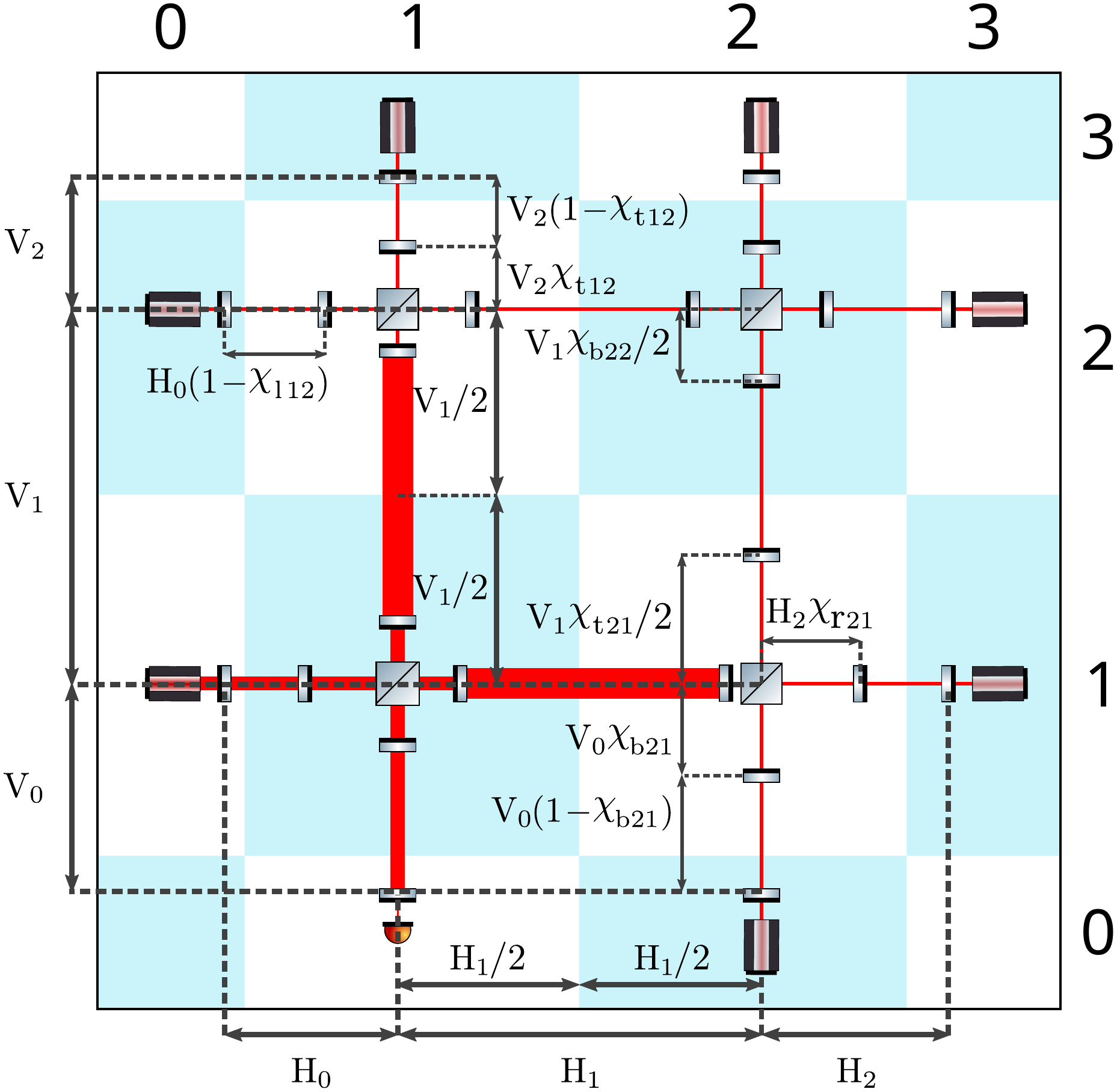}
    \caption{\textbf{Positioning of the optical elements.} To keep the grid structure, we locate the mirrors of the optical path based on fractions of larger distances, vertical $\{V_i\}$ and horizontal $\{H_j\}$, which are also fractions of a maximum vertical and horizontal size, up to 4km. For the sake of readability, we keep the mirrors within the cells, so the mirrors between beam splitters can be localized as a fraction of half the gap between beam splitters. The distance between light sources and their nearest mirrors is fixed, since that does not affect the design performance. Same applies for the detector and its nearest mirror.}
    \label{fig:distancesUIFO}
\end{figure}

In the UIFO framework, the optical components are located within an irregular rectangular grid (see Figure \ref{fig:distancesUIFO}). To locate each element, we set the distances between columns and rows of beam splitters and then locate the mirrors in between based on fractions of the largest distances. We fix the distances between the outer mirrors and their closest light source (or detector), since they do not affect the GWD performance. The values of the vertical and horizontal distances, $V_i$ and $H_j$, are also fractions of a larger tunable length, which can take up to 4km.

With the fraction based location the mirrors could overlap. To prevent that nonphysical situation, we impose a minimum value $\epsilon$ to all the distances described. This works because the number of elements between beam splitters is always the same, and therefore the number of distances. If we wanted to remove elements we should be cautious with these fixed gaps. 

\subsection{Discrete design choices}
Having set the UIFO topology and distances there are a few discrete design choices to make, like where to locate the detector, or which type of light sources to use. We also have to chose if we place beam splitters or Faraday isolators in the grid intersections, and how to orient them. We could also decide, for each optical component, whether we want it to be there at all.

In our experiments, we always place the detector in the same grid cell and only use lasers as light sources, not squeezers. We do not use Faraday isolators either and the orientation of the beam splitters is always the same. We are keeping it simple for now.

\subsection{Optical components}
Except for the detector, all optical components have, at least, one continuous parameter to tune. Table \ref{tab:components} summarizes the range of parameter values for the UIFO designs, before they are normalized (from 0 to 1) for training. Finesse can also simulate other elements and parameters, all listed and documented \cite{finesse}.

It is worth noticing that when you `turn off' a light source, both for squeezers and lasers, by setting their parameters to 0, their effect on the system ceases. That is not the case for mirrors and beam splitters. We can never ignore them because they have a minimum loss, and because they are still susceptible to power damages. To properly `turn them off' we must remove them completely.

\begin{table}[h!]
\centering
\begin{tabular}{@{}llll@{}}
\toprule
\textbf{Components}                                                            & \textbf{Parameters} & \textbf{Min} & \textbf{Max} \\ \midrule
\multirow{2}{*}{\begin{tabular}[c]{@{}l@{}}Mirrors and\\ beam splitters\end{tabular}} & Reflectivity       & 0            & 0.99998      \\
                           & Phase ($^\text{o}$)   & 0    & 360 \\ \midrule
Lasers                     & Power (W)   & 0    & 200 \\ \midrule
\multirow{2}{*}{Squeezers} & Factor (db) & 0    & 20  \\
                           & Angle ($^\text{o}$)   & 0    & 360 \\ \bottomrule
\end{tabular}
\caption{Range of values (before normalization) for different optical components. In Finesse, the mass of mirrors and beam splitters is infinite by default, we leave it like that in all our experiments. Mirrors and beam splitters have a minimum loss and transmission of 5ppm and 15ppm, respectively, which limits their reflectivity.}
\label{tab:components}
\end{table}

\section{Datasets} \label{app:datasets}
The 1 million training samples and 100K testing samples are stored in H5 data files with 1000 pairs of inputs-outputs each, normalized from 0 to 1. The inputs are the linearly normalized parameters of distances (lengths up to 4km and fractions of these) and optical components (described in Table \ref{tab:components}). The loss and masses of mirrors and beamsplitters are not included as they are constant.

The outputs are the sensitivity values of 101 frequency values (a geometric sequence from 1Hz to 10kHz), and the light power acting on the detector, mirrors, and beam splitters. As with the loss and masses, we do not include the frequency values in the dataset, as they are always the same. Since both sensitivities and powers range across several orders of magnitude, we must not rescale them directly from 0 to 1. Instead, and only during training/testing, we rescale the logarithm of their values
\begin{equation}
    \mathcal{N}(x) \equiv \frac{\log_{10}(x + \epsilon) + b}{S},
\end{equation}
where the values of $e$, $b$, and $S$ are (0, 30, 30) for the sensitivities, and ($10^{-6}$, 6, 16) for the powers. These choices are motivated by the expected range of sensitivities, usually from $10^{-25}$ to $10^{-10}$ (by definition, never 0 or higher than 1), and the expected range of powers, from 0 to $10^7$ (beyond that, everything burns). 

\section{Surrogate model architecture}\label{app:architecture}

\begin{figure*}[h!]
    \centering
    \begin{subfigure}[b]{0.33\textwidth}
        \centering
        \includegraphics[width=\textwidth]{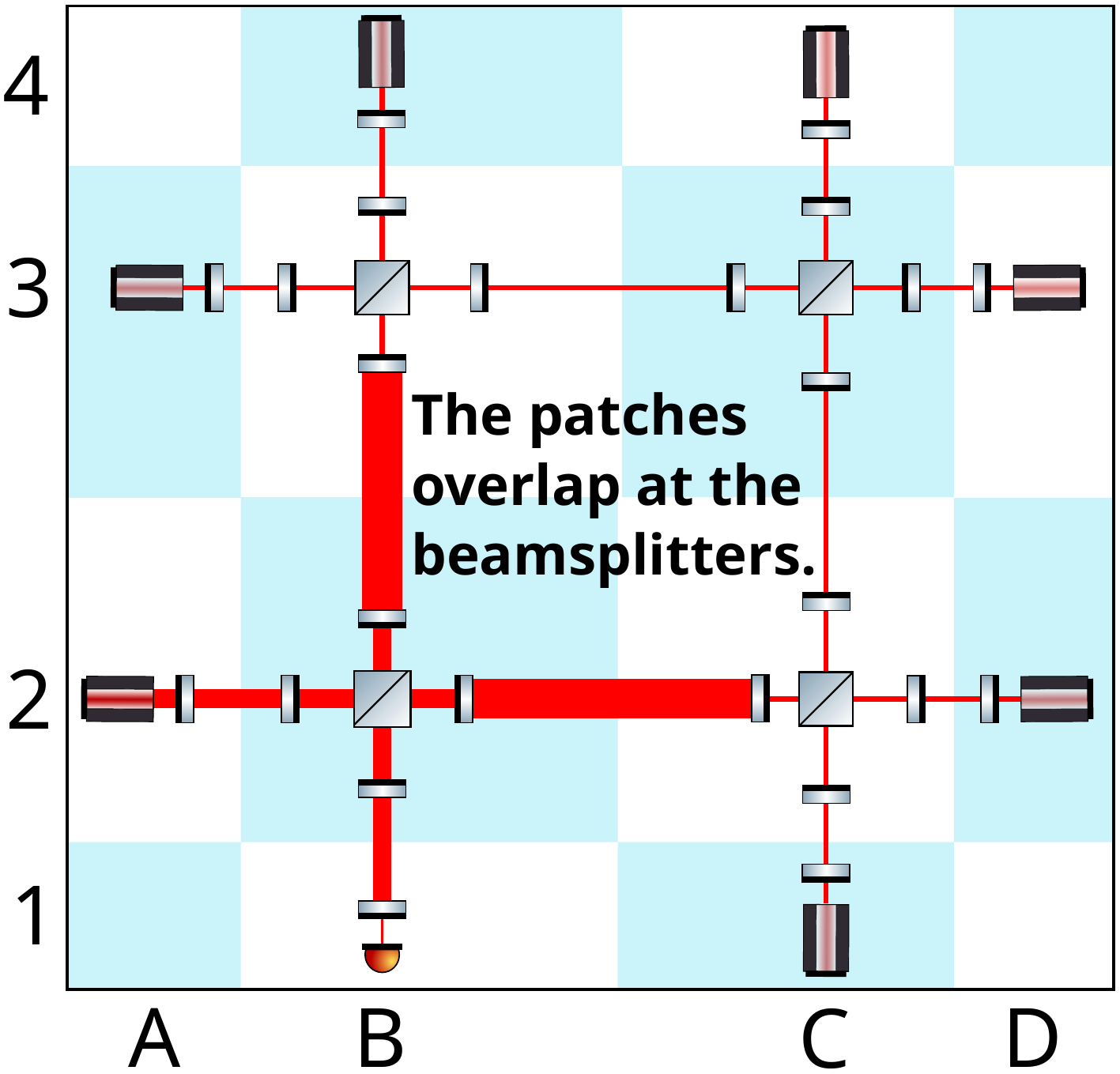}
        \caption{Original grid representation.}
        \label{fig:patch1}
    \end{subfigure}
    \hfill
    \begin{subfigure}[b]{0.32\textwidth}
        \centering
        \includegraphics[width=\textwidth]{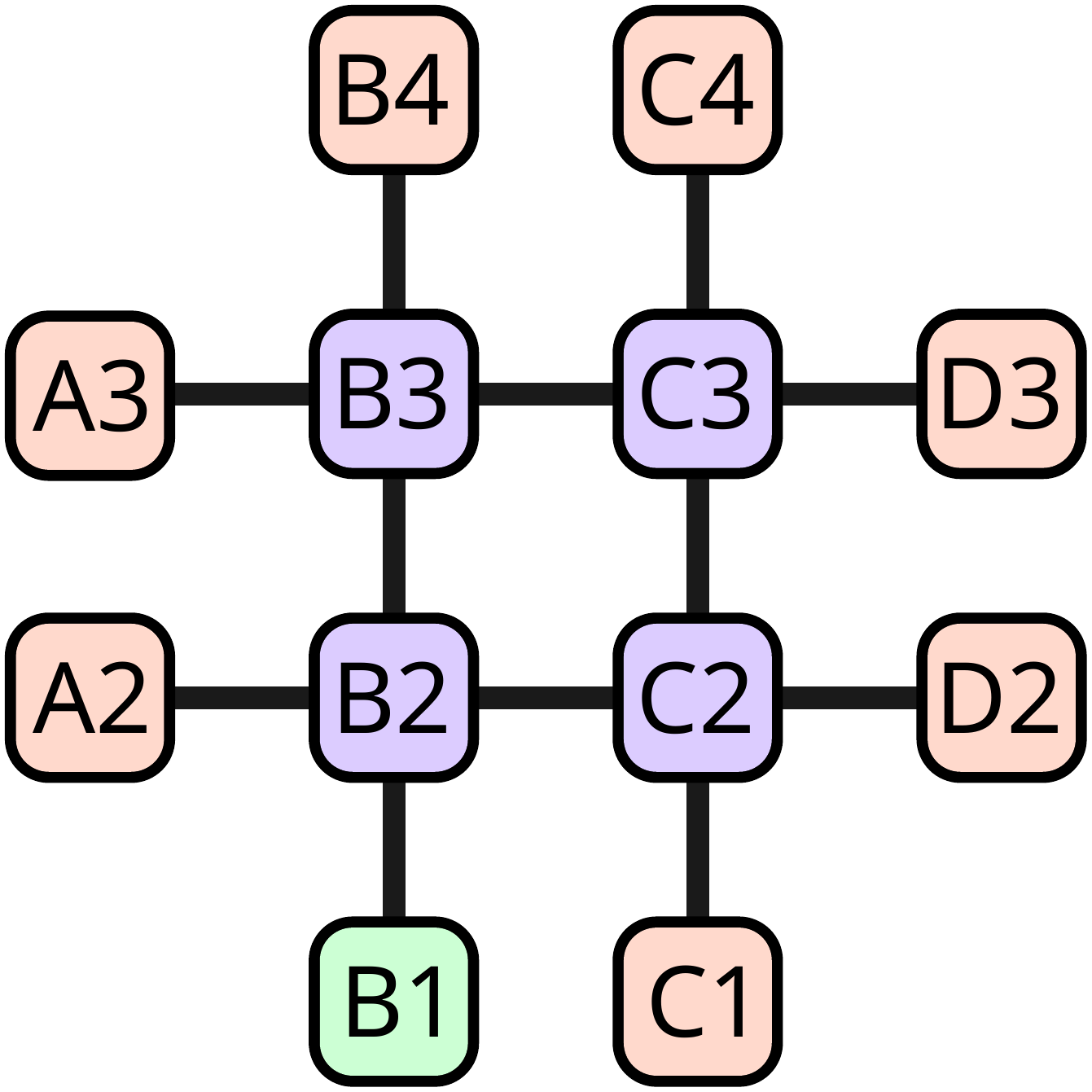}
        \caption{Heterogeneous attributed graph.}
        \label{fig:patch2}
    \end{subfigure}
    \hfill
    \begin{subfigure}[b]{0.32\textwidth}
        \centering
        \includegraphics[width=\textwidth]{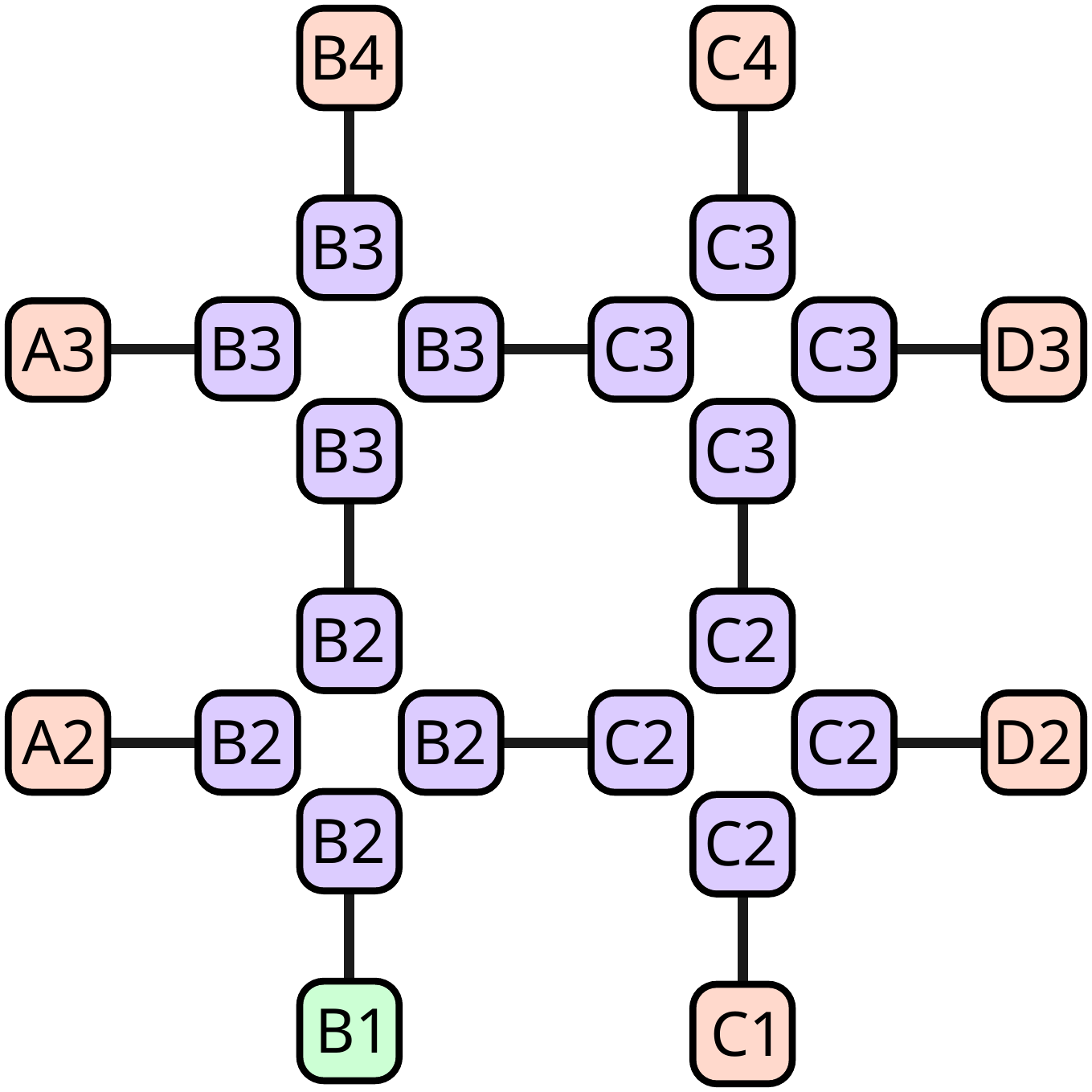}
        \caption{Division into subgraphs/patches.}
        \label{fig:patch3}
    \end{subfigure}
    \caption{\textbf{Patch extraction from the gravitational wave detector.} We encoded GWD into rectangular grids \textbf{(a)}, which can be seen as graphs with different types of nodes, and multiple attributes at nodes and edges \textbf{(b)}. Dividing the rectangular graph into overlapping subgraphs, we group the parameters that characterize them into patches \textbf{(c)}. The three types of nodes, light sources (orange), beam splitters (purple), and detector (green), while the mirror properties are encoded into the edges that connect the nodes. Since the patches have different number of attributes we pad the patches up to a common length. For simplicity, we group the vertical and horizontal distances, $\{V_i\}$ and $\{H_j\}$, into 2 additional patches.} 
    \label{fig:patches}
\end{figure*}

To mimic the behavior of the Finesse simulator we trained a transformer-based surrogate model grids \cite{mixers4images,mixers4graphs}. The parameters of neighbor elements were merged into tokens. We divided the grid by the beamsplitters (see Figure \ref{fig:patches}) and add their parameters to the four neighboring subgraphs. These overlapping patches were projected into a higher-dimensional space via a linear layer followed by a sinusoidal activation. This transformation, also known as Fourier feature mapping \cite{fourierfeatures}, proved essential for capturing the high-frequency features of the GWD (see Figure \ref{fig:UIFO}), overcoming the spectral bias of standard neural networks \cite{frequencybias,spectralbias}. The tokens are then processed by 8 transformer layers. After processing the tokens, they were projected into a lower dimension and concatenated. With a final multilayer perceptron, we obtain an array of predicted sensitivities and optical powers.

Since the transformer encoder is invariant under permutation of patches, we add 2 trainable positional encoding to the patches: before and after they are processed by the Fourier mapping.

\section{GPU resources used for surrogate-based inverse design} \label{app:resources}
The training times showed in Table \ref{tab:massivetable} are averaged from the data shown in Table \ref{tab:trainingtimes}. The inverse design of 20K samples over 16384 optimization steps took 16, 40 and 85 minutes on average, for the 3 grid sizes. We used \textit{AMD Instinct MI300A APU} units.
\begin{table}[h!]
\centering
\begin{tabular}{|c|c|ccc|}
\hline
\multirow{2}{*}{\textbf{\begin{tabular}[c]{@{}c@{}}Generation\vspace{-1mm}\\of surrogate\end{tabular}}} &
  \multirow{2}{*}{\textbf{Seed}} &
  \multicolumn{3}{c|}{\textbf{UIFO size (\# parameters)}} \\ \cline{3-5} 
                     &   & \multicolumn{1}{c|}{\textbf{1x1 (29)}} & \multicolumn{1}{c|}{\textbf{2x2 (85)}} & \textbf{3x3 (169)} \\ \hline
\multirow{2}{*}{0th} & 0 & \multicolumn{1}{c|}{20' 12''} & \multicolumn{1}{c|}{27' 00''} & 47' 00''               \\ \cline{2-5} 
                     & 1 & \multicolumn{1}{c|}{20' 05''} & \multicolumn{1}{c|}{31' 23''} & 40' 57''           \\ \hline
\multirow{2}{*}{1st} & 0 & \multicolumn{1}{c|}{45' 09''} & \multicolumn{1}{c|}{56' 04''} & 1h 30' 32''        \\ \cline{2-5} 
                     & 1 & \multicolumn{1}{c|}{44' 29''} & \multicolumn{1}{c|}{56' 44''} & 1h 13' 20''        \\ \hline
\multirow{2}{*}{2nd} & 0 & \multicolumn{1}{c|}{1h 07' 12''} & \multicolumn{1}{c|}{1h 36' 27''} & 2h 25' 46''        \\ \cline{2-5} 
                     & 1 & \multicolumn{1}{c|}{1h 15' 58''} & \multicolumn{1}{c|}{1h 47' 48''} & 1h 39' 36''        \\ \hline
\multirow{2}{*}{3rd} & 0 & \multicolumn{1}{c|}{1h 42' 31''} & \multicolumn{1}{c|}{2h 04' 59''} & 3h 06' 26''        \\ \cline{2-5} 
                     & 1 & \multicolumn{1}{c|}{1h 43' 32''} & \multicolumn{1}{c|}{2h 25' 05''} & 3h 12' 27''        \\ \hline
\multirow{2}{*}{4th} & 0 & \multicolumn{1}{c|}{2h 20' 13''} & \multicolumn{1}{c|}{2h 34' 05''} & 4h 09' 31''        \\ \cline{2-5} 
                     & 1 & \multicolumn{1}{c|}{2h 12' 22''} & \multicolumn{1}{c|}{2h 48' 26''} & 3h 25' 32''        \\ \hline
\multirow{2}{*}{\textbf{\begin{tabular}[c]{@{}c@{}}Total\vspace{-1mm}\\time\end{tabular}}} &
  0 &
  \multicolumn{1}{c|}{6h 15' 17''} &
  \multicolumn{1}{c|}{7h 38' 35''} &
  11h 59' 15'' \\ \cline{2-5} 
                     & 1 & \multicolumn{1}{c|}{6h 16' 26''}                  & \multicolumn{1}{c|}{8h 29' 26''}        & 10h 11' 52''       \\ \hline
\end{tabular}
\caption{\textbf{Training times of the surrogate models for the 3 UIFO sizes.}}
\label{tab:trainingtimes}
\end{table}

\section{Optimization with Finesse}
\begin{figure}[h!]
  \centering
  \includegraphics[width=\textwidth]{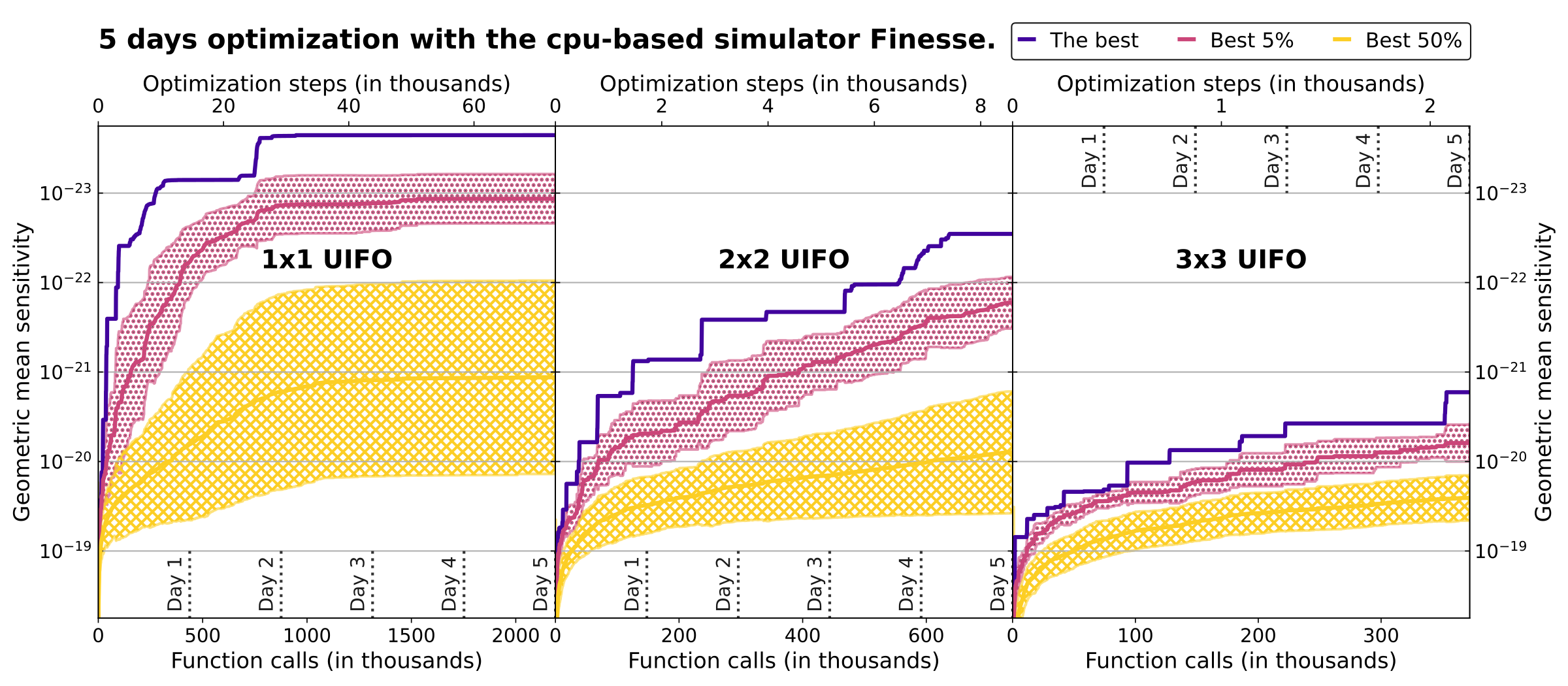}
  \caption{\textbf{Optimization of 256 designs with Finesse during 5 days.} Using the Adam optimizer with annealed noise (same than for the surrogates) we optimize 256 independent input parameter sets, randomly initialized, to minimize the loss from Eq.\eqref{eq:loss}. Lacking automatic differentiation, we must use finite differences to compute the gradients, taking 30, 86, and 170 evaluations per gradient for the different UIFO sizes: 1x1, 2x2, and 3x3. With approximate running times of 0.2, 0.6, and 1.2 seconds respectively, each optimization step might take more than 3 minutes for the largest system. The plotted data considers the best sensitivity ever achieved by the samples at a given optimization step, ignoring the decreases from the gradient computations and/or the optimization process.}
  \label{fig:direct_optimization}
\end{figure}

Figure \ref{fig:direct_optimization} shows the detailed results of the design optimization run with Finesse for 5 days in 256 parallel jobs for each size of the UIFO.

\end{document}